\pdfoutput=1

\documentclass[11pt]{article}

\usepackage[preprint]{acl}

\usepackage{times}
\usepackage{latexsym}
\usepackage{booktabs}
\usepackage{float}
\usepackage{enumitem}

\usepackage[T1]{fontenc}

\usepackage[utf8]{inputenc}

\usepackage{microtype}

\usepackage{inconsolata}

\usepackage{graphicx}

\usepackage{amsmath}

\DeclareMathOperator*{\argmax}{arg\,max}

\title{An Explainable Framework for Misinformation Identification via Critical Question Answering}

\author{Ramon Ruiz-Dolz \and John Lawrence\\
  Centre for Argument Technology (ARG-tech) \\ 
  University of Dundee \\
  United Kingdom \\
  \texttt{\{rruizdolz001,j.lawrence\}@dundee.ac.uk}}

\begin{document}
\maketitle

\begin{abstract}

    Natural language misinformation detection approaches have been, to date, largely dependent on sequence classification methods, producing opaque systems in which the reasons behind classification as misinformation are unclear. While an effort has been made in the area of automated fact-checking to propose explainable approaches to the problem, this is not the case for automated reason-checking systems. In this paper, we propose a new explainable framework for both factual and rational misinformation detection based on the theory of Argumentation Schemes and Critical Questions. For that purpose, we create and release \textsc{NLAS-CQ}, the first corpus combining 3,566 textbook-like natural language argumentation scheme instances and 4,687 corresponding answers to critical questions related to these arguments. On the basis of this corpus, we implement and validate our new framework which combines classification with question answering to analyse arguments in search of misinformation, and provides the explanations in form of critical questions to the human user.
\end{abstract}

\section{Introduction}

In recent years, the spread of misinformation has become a major societal concern \cite{misinformation2023ipsos} due to a range of factors, including the widespread growth of online social networking sites \cite{naeem2021exploration} and popularisation of generative large language models \cite{jiang2023disinformation}, together with a lack of consolidated countermeasures \cite{gausen2021can}. Misinformation can be identified at two different levels; factual misinformation and rational misinformation \cite{visser2020reason}. The former involves the use of factually false claims (e.g., \textit{fake news}), while the latter involves the use of invalid reasoning (e.g., \textit{fallacies}). It is also possible to observe this dichotomy in the research carried out in Artificial Intelligence (AI) and Natural Language Processing (NLP), where works aimed at automatic misinformation identification are clearly oriented towards either automated fact-checking (i.e., factually false information) \cite{guo2022survey} or the automatic identification of fallacies (i.e., invalid reasoning) \cite{jin2022logical}.

While automated fact-checking is largely based around information retrieval and validating pieces of information against \textit{reliable} sources, automated fallacy identification requires further analysis of the reasoning behind the natural language that makes the argument, in addition to natural language modelling itself \cite{visser2020reason}. Given that a fallacy is caused by a fault in the underlying reasoning of an argument, a model capable of capturing the logical structure of the argument in addition to its natural language is needed \cite{ruiz2023detecting}.

The Argumentation Scheme model of human argumentation proposed by \cite{walton2008argumentation} provides a framework to effectively model more than 60 stereotyped patterns of argumentative reasoning (i.e., argumentation schemes), and a wide set of critical questions to challenge their validity from both factual and logical viewpoints. 
With this model of argumentation, and specifically the critical questions, we can address both factual and logical issues in natural language arguments containing misinformation. Furthermore, the critical questions themselves provide us with a much-needed \textit{explainability} in the development of such systems, which is of utmost importance in the development of critical thinking that enables the human users to question and detect pieces of misinformation without having to depend on an automated system.

Let us consider the \textit{Argument from Position to Know} scheme as an example, \cite{walton2008argumentation} define this argument as:


\begin{itemize}[leftmargin=*]
    \item[] \underline{\textit{Major Premise}}: Source $s$ is in position to know about things in a certain subject domain $f$ containing proposition $p$.
    \item[] \underline{\textit{Minor Premise}}: $s$ asserts that $p$ is true (false).
    \item[] \underline{\textit{Conclusion}}: $p$ is true (false).
\end{itemize}


An argumentation scheme thus provides a set of abstract variables (i.e., $s$, $f$, and $p$ in this case) that can be replaced with natural language text, together with the connections (in natural language) between these variables required to make a specific argumentation scheme instance. Furthermore, Walton provides the set of predefined critical questions, the answers to which determine the validity of the argument from both factual and logical viewpoints:


\begin{itemize}[leftmargin=*]
    \item[] \underline{\textit{CQ1}}: Is $s$ in position to know whether $p$ is true?
    \item[] \underline{\textit{CQ2}}: Is $s$ an honest/trustworthy/reliable source?
    \item[] \underline{\textit{CQ3}}: Did $s$ assert that $p$ is true?
\end{itemize}


Being unable to provide a \textit{good}\footnote{A good answer is one that supports the soundness of the argument.} answer for all the critical questions will result in a potential piece of misinformation, and we will know the exact reasons why it can be considered as such. In the running example, either if $s$ is not in position to know about $p$, $s$ is not a reliable source, or $s$ has not asserted that $p$ is true, as it is said in the argument, its credibility will be undermined. Therefore, combining argumentation schemes with NLP techniques allow us to approach the problem of automatically identifying misinformation in a manner which is both theoretically well grounded, and transparent.

In this paper we propose a new explainable framework for misinformation detection by automatically answering the critical questions, specifically addressing the questions of argument scheme classification and critical question answering. For that purpose, we create and publicly release \textsc{NLAS-CQ}, the first corpus containing 3,566 textbook-like natural language argumentation schemes in English where 1,794 of them have been annotated with answers to their respective critical questions. On the basis of this corpus, we propose our new framework which combines classification with question answering to analyse arguments in search of misinformation. Compared to most previous work, in which the detection of natural language fallacies is approached as a natural language sequence classification problem \cite{da2019fine}, \cite{sahai2021breaking}, \cite{jin2022logical}, in our proposed framework we rely on concepts of argumentation theory to complement the NLP sequence modelling algorithms. For that purpose, we first classify the natural language sequence (i.e., an argument) into one of the different argumentation scheme classes. Once the reasoning pattern behind an argument has been identified, we have access to the set of critical questions that help us assess its validity according to the framework proposed by Walton \cite{walton2008argumentation}. The final step in our proposed framework consists automatically answering the critical questions, providing us with additional information to detect a potential misinformation. Our contribution is therefore threefold:

\begin{itemize}
    \item[(i)] We release \textsc{NLAS-CQ}, the first corpus containing 3,566 textbook-like natural language argumentation scheme instances in English along with 4,687 corresponding answers to critical questions related to these scheme instances.
    \item[(ii)] We propose the first explainable framework for misinformation detection in natural language arguments. This framework allows not only to identify a potential piece of misinformation, but also the reasons why the argument can be considered as such.
    \item [(iii)] We establish baselines in the new task of critical question answering, an explainable and human-centered approach to the automatic identification of misinformation from both factual and rational viewpoints.
\end{itemize}  

\section{Related Work}

Misinformation has been studied by philosophers for many years \cite{fox1983information}, \cite{kuklinski2000misinformation}, however, it is only recently that we have become aware of the threat it represents in technologically developed societies with open access to unfiltered online media. Due to the ease with which information can be disseminated in online platforms \cite{fernandez2018online}, identifying pieces of misinformation has become a major challenge in the areas of computer science and NLP \cite{musi2023developing2}. A platform to track the spread of online misinformation was proposed by \cite{shao2016hoaxy}, which can be helpful to fight against its quick dissemination. Following the relevance of this topic, multiple studies have analysed the properties of online misinformation in recent years, in different domains such as the political \cite{shao2018anatomy}, climate change \cite{treen2020online}, or the recent pandemic \cite{musi2022developing} among others. All these studies converge on the fact that fact-checking and argumentative analysis of potential pieces of misinformation are a cornerstone of the fight against misinformation. 

Argumentative analysis or \textit{reason-checking} \cite{visser2020reason} is defined as an additional step in the process of identifying misinformation where, in addition to fact-checking, the reasoning structures and patterns are also analysed. \cite{musi2022fallacies} studies the links between the use of argumentative fallacies and the spread of misinformation in social media. This work is extended in a recent study \cite{musi2023developing}, where the authors present two chat-bots that make use of the concept of critical questions from \cite{walton2008argumentation} to guide the participants through an argumentative reasoning of information in a controlled experimental setup. Automated \textit{reason-checking} systems, however, have not been investigated in-depth yet, mainly due to the limited availability of data to train predictive models which is caused by the complexity of human argumentative analysis.

Regarding the development of automatic misinformation identification systems, most of the literature focuses on either binary classification tasks \cite{yu2017convolutional}, \cite{ijcai2024p723}, \cite{ijcai2024p281} or fact-checking \cite{guo2022survey}. A small number of works have focused on \textit{reason-checking} (e.g., identification of fallacies), releasing small corpora \cite{da2019fine}, \cite{sahai2021breaking}, and reporting initial results on fallacy classification tasks \cite{alhindi-etal-2022-multitask}, \cite{jin2022logical}. These approaches, however, address the misinformation identification problem as a sequence classification, relying on probabilistic models of natural language. As pointed out in recent work \cite{ruiz2023detecting}, modelling this problem as sequence classification without considering concepts from argumentation theory presents fundamental limitations, resulting in models that can not generalise well and present a significant number of false positives. This issue has a direct negative impact on the usefulness of such systems, labelling as a fallacy something that isn't just because it shares vocabulary with what in the training data has been annotated as a fallacy.

This limitation is closely related to an important aspect of automatic misinformation identification, explainability \cite{mishima2022survey}. Humans are a fundamental part of the process, as they are the intended \textit{receivers} of the outcomes of such systems. Thus it is fundamental to design them, and their underlying algorithms, in a way that can be easily accepted and understood by the final human users \cite{ijcai2022p706}. Work on automatic misinformation identification focused on fact-checking (e.g., fake news detection) has been actively investigating ways of improving the explainability of the proposed approaches, for example by including knowledge bases as part of the process and extracting the facts that support the final prediction \cite{shu2019defend}, \cite{ijcai2021p646}. This aspect, however, remains unexplored from the \textit{reason-checking} perspective.

This paper, therefore, advances the literature in all the aforementioned limitations by proposing an explainable framework that allows to automatically identify potential pieces of misinformation considering both factual and rational aspects, and provides the user with the set of reasons (i.e., critical questions) behind the output, not only preventing its dissemination but also supporting the development of critical thinking skills.


\section{Framework}
\label{framework}

Our proposed explainable framework consists of a pipeline with two main components: the Argumentation Scheme Classification (ASC) module and the Critical Question Answering (CQA) module. Different to previous work in automatic misinformation identification, where only sequence modelling is considered to determine whether a natural language sequence is a potential piece of misinformation or not, in our work, we first determine the reasoning pattern (i.e., argumentation scheme) that a natural language sequence follows, and then question its validity by answering the critical questions. In this way, the ASC enriches the analysis with additional argumentative information, and the CQA estimates the most likely answers to the critical questions, providing us with relevant information of the strength and the soundness of an argument from both factual and rational viewpoints, as well as the reasons on which these estimates are based. 

\begin{figure}
    \centering
    \includegraphics[width=0.7\columnwidth]{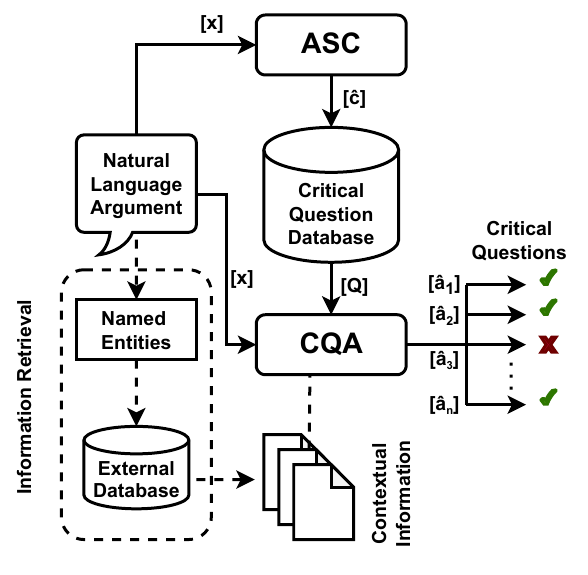}
    \caption{Critical Question-based Explainable Framework for Misinformation Detection.}
    \label{fig:framework}
\end{figure}

As depicted in Figure \ref{fig:framework}, where the overall architecture of the proposed framework is represented, the ASC module estimates the argumentation scheme a given argument belongs to by modelling the following conditional probability,


\begin{equation}
    \hat{c} = \argmax_{c \in C} P(c|x)
    \label{eq:1}
\end{equation}

where the correct argumentation scheme class $\hat{c}$ for a given argument $x$ is the one that maximises the conditional probability of that argument $x$ belonging to that class $\hat{c}$ from the complete set of possible argumentation scheme classes $c \in C$. The output of the ASC module determines the set of critical questions $Q$ to be answered by the CQA module, which provides an answer for each of these critical questions. The estimations produced by the CQA module are the result of modelling the following conditional probability distribution,


\begin{equation}
    \hat{a} = \argmax_{a \in A} P(a| q, x)
    \label{eq:2}
\end{equation}

where the correct answer $\hat{a}$ corresponds to the answer $a$ in the space of possible answers $A$ (i.e., yes or no) that maximises the conditional probability for a given (critical) question $q$ (from the complete set of questions $Q$ related to scheme $\hat{c}$), and an argument $x$. Contextual information resulting from an information retrieval process can also be used to enrich the natural language information included in the argument $x$. For the sake of narrowing down the focus of this paper, however, in the subsequent sections we will assume that the contextual information has already been retrieved, and it is available for complementing the information provided to the CQA module as an input.

Let us illustrate the proposed explainable framework with an example. Having our natural language argument input (i.e., $x$) defined as follows,

\begin{itemize}
    \item[] ``\textit{Generally, driving a car requires a license and adherence to safety regulations, which is similar to owning a gun. It is important for individuals to pass a background check and be required to receive proper training before owning a gun to ensure safety. Therefore, it is reasonable for gun ownership to require passing a background check and receiving proper training, similar to driving a car.}''
\end{itemize}

the first step will be to process the argument with the ASC identifying the argumentation scheme to which it belongs, in this case, an \textit{Argument from Analogy} (i.e., $\hat{c}$). Once the reasoning structure underlying our argumentative input has been identified, we can retrieve the set of associated critical questions as defined in \cite{walton2008argumentation}. In this case, we will have three different critical questions that challenge the validity of our \textit{Argument from Analogy},


\begin{itemize}[leftmargin=*]
\setlength\itemsep{-0.3em}
    \item[] \textit{CQ1:} Are there differences between the two cases that would tend to undermine the force of the similarity cited?
    \item[] \textit{CQ2:} Is the claim also true in the first case?
    \item[] \textit{CQ3:} Is there some other case that is also similar to the first case but in which the claim is false?
\end{itemize}


The CQA module will be in charge of automatically answering these questions. In this case, \textit{CQ1} and \textit{CQ2} (i.e, $q_1$ and $q_2$) will be answered affirmatively (i.e., $\hat{a_1}$ and $\hat{a_2}$), while \textit{CQ3} (i.e., $q_3$) will be answered negatively (i.e, $\hat{a_3}$). We can observe how, in this example, the first critical question does not have a satisfactory response, undermining the validity of the argument, and making it a potential piece of misinformation. The explanation supporting this statement can be directly inferred from the own critical question, which is that driving a car is a significantly different case than owning a gun.

\section{The \textsc{NLAS-CQ} Corpus}


To implement and evaluate the proposed explainable framework, we developed the first publicly available corpus of Natural Language Argumentation Schemes and Critical Questions (\textsc{NLAS-CQ}\footnote{The corpus is accessible upon request under a CC BY-NC-SA 4.0 license.}). We created the \textsc{NLAS-CQ} corpus by combining the use of generative LLMs and human annotation. It is important to note that all the text generated by LLMs included in our corpus was validated by annotators with extensive experience in argumentation. The corpus was therefore created in two phases, first, the generation of NLAS resulting in a large collection of textbook-like natural language arguments, and second, the annotation of CQs producing the first publicly available collection of in-depth analysed natural language argumentation schemes. 

\subsection{Argumentation Scheme Generation}

In the first phase, we followed a communication strategy with the API of \cite{openai2023gpt4}, applying prompting techniques to generate scheme instances. We aimed to maintain the original structure of each argument scheme proposed by \cite{walton2008argumentation}. To achieve this, the general structure of our prompt comprised three main parts: the argument type specification, the argumentation scheme structure, and the output format.

For the argument type specification, we request that the system generates a specific argument type (e.g., Argument from Position to Know) by defining the stance (i.e., in favour or against) and the topic (e.g., open source research). Following these guidelines (see the Technical Appendix for the detailed prompts), an example of the argument type specification would be: ``\textit{Provide a position to know argument in favour of open source research}''. We complemented the prompt with a template of the argumentation scheme structure to be generated (similar to the one presented in the Introduction). Finally, we specified the format in which we wanted the output. We requested a JSON-like output respecting the structure of the argumentation scheme and storing all the relevant information such as the scheme name, the topic and the stance. 

Therefore, in order to complete the first phase of natural language argumentation scheme generation, we prompted the model to generate 4,000 arguments, including a total of 20 argumentation schemes, 100 topics, and 2 stances per topic (see the Technical Appendix for a complete list of all the variables). Once the 4,000 argumentation scheme instances were generated, we proceeded with the human validation of this automatically generated data. For that purpose, five expert annotators were involved in the process. During the validation, the experts looked exclusively at the adequacy of the argument generated in relation to the pattern of each argumentation scheme, the topic, and the stance. Our automatically generated NLAS were labelled as valid or not valid based on these three aspects. Arguments that were classified as not valid (i.e., arguments that did not adhere to the argumentative structure, topic, or stance) were discarded. To validate the quality of the annotation in this first phase, we carried out an Inter Annotator Agreement (IAA) test with 10\% of the generated NLAS, reporting a 0.65 Cohen's Kappa \cite{cohen1960coefficient}, indicating substantial agreement between the expert annotators.

With this method, we were able to obtain 3,566 valid NLAS in English out of the 4,000 prompted argumentation scheme instances (i.e., 89.1\%) which allowed us to start the second phase of annotating the critical questions.

\subsection{Critical Question Annotation}

In the second phase, we annotated the critical question answers for the previously validated NLAS. Given the complexity of this task, some adjustments were made to the original critical questions in an attempt to simplify the annotation while keeping the focus on the factual and logical validity of the argument. First of all, we made sure that all the critical questions were binary answer questions, so that the possible answers were yes or no. For example, we adapted the first CQ of the argument from popular practice defined as ``\textit{What actions or other indications show that a large majority accepts A?}'' to the binary answer version of the same question: ``\textit{Are there actions or other indications that show that a large majority accepts A?}''. In this way, it was not only easier for the annotators to complete their assignments, but the data was also prepared for a more straightforward question answering task with only two possible answers. 

Therefore, the expert annotators read the arguments together with their critical questions and selected one of the following options: ``\textit{No.}'', ``\textit{Possibly no.}'', ``\textit{I'm not sure.}'', ``\textit{Possibly yes.}'', and ``\textit{Yes}'', depending on their level of confidence in their response. In addition to that, we also requested the annotators to provide the supporting evidence that justified their answers (i.e., the contextual information), which was compiled in an additional \textit{.csv} file (see the Technical Appendix for specific examples of contextual information included in our corpus). To validate the quality of this second phase of annotation, we carried out another IAA test with a 10\% of the samples, reporting a 0.42 Cohen's Kappa which represents a moderate agreement between the annotators \cite{landis1977measurement}. For the IAA test we considered both ``\textit{No.}'' and ``\textit{Possibly no.}'', and ``\textit{Possibly yes.}'' and ``\textit{Yes}'' as an agreement between annotators.

In the end, 10 different argumentation schemes were completely annotated with their critical questions (see the Technical Appendix for more detail on the data distribution), making it a total of 1,794 NLAS with their respective 4,687 CQs. The \textsc{NLAS-CQ} represents not only the largest publicly available corpus of natural language argumentation scheme instances in English consisting of a total of 3,566 NLAS, but it is also the first publicly available corpus containing answers to the critical questions. It is possible this way to provide insight into the validity of arguments from both factual and rational points of view. All this makes \textsc{NLAS-CQ} a valuable resource for the NLP and argumentation communities, allowing for new experimental paradigms to start exploring challenging tasks such as the automatic detection of misinformation beyond fact-checking.

\section{Experiments}

\subsection{Experimental Setup}

The experimentation done in this paper was carried out at different levels. Aiming at modelling the conditional probability defined in Equation \ref{eq:1}, and to implement the ASC module, we tried two different approaches. The first consisted of training a Support Vector Machine (SVM) \cite{hearst1998support} classifier using the sentence embedding representations \cite{reimers2019sentence} produced by a Transformer \cite{vaswani2017attention} language model. The second consisted of fine-tuning a \textit{RoBERTa-large} \cite{liu2019roberta} model by applying \textit{Inductive Transfer Learning} \cite{torrey2010transfer} on our corpus. Both models were selected based on the size of the training corpus, and in previous results in similar NLP tasks \cite{ruiz2021transformer}.

To implement the CQA module we modelled the conditional probability in Equation \ref{eq:2} with two different approaches. First, we fine-tuned a \textit{RoBERTa-large} model on a Binary Question Answering task \cite{ghosal-etal-2022-two}, where the natural language input contained the argument, the contextual information (i.e., the supporting evidence provided by the annotators) if added, and the question, which could only be answered affirmatively or negatively. 
In our second approach, we used a similar input to prompt both Mixtral-8x7B-Instruct-v0.1 \cite{jiang2024mixtral} and GPT-4 \cite{openai2023gpt4} to answer the critical question via Generative Question Answering \cite{yin2016neural} as follows (see the Technical Appendix for the detailed specifications of the prompts used):


\begin{itemize}
\setlength\itemsep{-0.3em}
    \item[] ``\textit{context: [NLAS]+[CI] }
    
    \item[] \textit{question: [CQ] }
    
    \item[] \textit{answer:} ''
\end{itemize}


Regarding hyperparameters, for the SVM we used an \textit{rbf} kernel with a $C = 100$ and a scaled gamma. With respect to the RoBERTa architectures on both ASC and CQA tasks, we set up a learning rate of 1e-5, with a batch size of 32, and a weight decay of 0.1, and fine-tuned the pre-trained model for 5 epochs in each of the two tasks.

With respect to the evaluation metrics, we used \textit{Precision}, \textit{Recall}, and macro averaged \textit{F1-score}, being suitable and informative metrics for multi-class and imbalanced classification tasks \cite{godbole2004discriminative}. All the previously defined experiments have been implemented and run in an RTX 3090 with 24Gb of VRAM, 32Gb of RAM, and an Intel Core i7-9700k CPU machine.

\begin{table}
    \centering
    \resizebox{0.9\columnwidth}{!}{%
    \begin{tabular}{c c c c c}
    \toprule
    \textbf{Task (corpus)} & \textbf{Unit} & \textbf{Train} & \textbf{Development} & \textbf{Test}\\ \midrule\midrule
    ASC (\textsc{NLAS-CQ}) & NLAS & 2,852 & 357 & 357 \\ \midrule 
    ASC (QT30) & Inference & - & 187 & 100 \\ \midrule 
    CQA (\textsc{NLAS-CQ}) & CQ & 2,963 & 443 & 462 \\ \bottomrule
    \end{tabular}}
    \caption{Distribution of the data used in the experiments for Argument Scheme Classification and Critical Question Answering.}
    \label{tab:data}
\end{table}

In order to provide consistent results of the evaluation of our ASC and CQA modules in the misinformation detection task, we used the \textsc{NLAS-CQ} corpus containing 3,566 \textsc{NLAS} and 4,687 CQs. We divided this corpus into \textit{train}, \textit{development}, and \textit{test} following an 80-10-10 distribution. Furthermore, to validate the performance of the pre-trained models on the texbook-like arguments of the \textsc{NLAS-CQ} corpus, we also included a small validation dataset containing the same annotated argumentation schemes in real dialogue arguments extracted from the QT30 corpus \cite{hautli2022qt30}. This validation dataset contained 287 natural language argumentation schemes and was divided into development (i.e., fine-tuning) and test according to a 65-35 distribution. Table \ref{tab:data} contains the details of how the samples were distributed between the different data splits. A sample in ASC is represented by an individual NLAS, while in CQA each sample is represented by an individual critical question associated to each NLAS. For this reason, there are more samples in the data for the CQA task, since an NLAS has, at least, two associated critical questions. 

\subsection{Results}


\subsubsection{Argumentation Scheme Classification (ASC)}

\begin{table}
    \centering
    \resizebox{0.8\columnwidth}{!}{%
    \begin{tabular}{l c c c}
    \toprule
    \textbf{Model} &  \textbf{Precision} & \textbf{Recall} & \textbf{F1-score}\\ \midrule
    Random Baseline & 4.6$\pm$0.94 & 4.8$\pm$1.03 & 4.6$\pm$0.97 \\ 
    eSVM-\textsc{asc} & 83.7$\pm$0.00 & 83.0$\pm$0.00 & 83.0$\pm$0.00 \\ 
    \textbf{RoBERTa-\textsc{asc}} & \textbf{99.2$\pm$0.20} & \textbf{99.2$\pm$0.20} & \textbf{99.2$\pm$0.20} \\ \bottomrule
    \end{tabular}}
    \caption{Results of the Argumentation Scheme Classification task pre-training on the \textsc{NLAS-CQ} corpus. The reported results have been averaged from 5 randomly initialised sequential runs.}
    \label{tab:asc}
\end{table}

Considering the size of the available corpora for argument scheme classification, two different approaches were considered for the ASC task. First, we trained an SVM (i.e., eSVM-\textsc{asc}) in a 20-class classification problem considering the schemes included in the \textsc{NLAS-CQ} corpus. Second, we fine-tuned a \textit{RoBERTa-large} model (i.e., RoBERTa-\textsc{asc}) on this same problem. The observed results after averaging 5 randomly initialised sequential runs considering these two approaches are depicted in Table \ref{tab:asc}. Given the nature of the \textsc{NLAS-CQ} corpus containing textbook-like arguments, we complemented our experiments with a second validation considering a small set of natural language argumentation schemes extracted from real dialogues belonging to the QT30 corpus \cite{hautli2022qt30}. In our validation experiments we included two different approaches, considering the 20 scheme classes (i.e., RoBERTa-\textsc{asc}), and grouping the schemes into 8 different scheme groups sharing similarities between their CQs as defined by \cite{walton2015classification} (i.e., RoBERTa-AF-\textsc{asc}). Furthermore, we also report the results of fine-tuning these models with the dialogue arguments (i.e., RoBERTa-\textsc{Dial}-\textsc{asc} and RoBERTa-\textsc{Dial}-AF-\textsc{asc}). The validation results have been summarised in Table \ref{tab:validation}. 

As we can observe, the performance in all, \textit{Precision}, \textit{Recall}, and \textit{F1-score} is very high when fine-tuning the RoBERTa model. This is due to the fact that different argumentation schemes have significant differences in their structure, which are directly reflected in the way they are instantiated in the \textsc{NLAS-CQ} texbook-like arguments. We can observe, however, how it becomes a significantly more difficult task when dealing with \textit{enthymematic} arguments used in real dialogues. However, by grouping the 20 schemes into the 8 groups and fine-tuning with a small dataset allowed us to keep the performance of the ASC systems, even when analysing arguments uttered in natural language dialogues with incomplete structures compared to the textbook-like arguments of \textsc{NLAS-CQ}.


\begin{table}
    \centering
    \resizebox{0.8\columnwidth}{!}{%
    \begin{tabular}{l c c c}
    \toprule
    \textbf{Model} &  \textbf{Precision} & \textbf{Recall} & \textbf{F1-score}\\ \midrule
    \textsc{RoBERTa-\textsc{asc}} & 0.8 & 4.5 & 1.0 \\ 
    \textsc{RoBERTa-AF-\textsc{asc}} & 45.5 & 38.1 & 31.7 \\  \midrule
    \textsc{RoBERTa-Dial-\textsc{asc}} & 1.5 & 8.6 & 2.4 \\ 
    \textbf{\textsc{RoBERTa-Dial-AF-\textsc{asc}}} & \textbf{58.8} & \textbf{48.9} & \textbf{44.8} \\   \bottomrule
    \end{tabular}}
    \caption{Results of the validation of our argumentation scheme identification models on the natural language dialogue schemes.}
    \label{tab:validation}
\end{table}

\subsubsection{Critical Question Answering (CQA)}

\begin{table}
    \centering
    \resizebox{0.95\columnwidth}{!}{%
    \begin{tabular}{l c c c}
    \toprule
    \textbf{Model} &  \textbf{Precision} & \textbf{Recall} & \textbf{F1-score}\\ \midrule
    RoBERTa-\textsc{cqa} & 70.9$\pm$0.77 & 71.6$\pm$1.23 & 70.7$\pm$0.54 \\ 
    Mixtral-8x7B & 30.3$\pm$0.00 & 30.7$\pm$0.00 & 28.7$\pm$0.00 \\ 
    GPT-4 & 64.8$\pm$0.00 & 63.2$\pm$0.08 & 63.5$\pm$0.08  \\ \midrule
    \textbf{RoBERTa-\textsc{cqa} [\textsc{ci}]} & \textbf{73.76$\pm$0.10} & \textbf{74.63$\pm$0.04} & \textbf{73.94$\pm$0.20} \\ 
    Mixtral-8x7B [\textsc{ci}] & 36.5$\pm$0.00 & 36.6$\pm$0.00 & 34.6$\pm$0.00 \\ 
    GPT-4 [\textsc{ci}] & 63.2$\pm$0.15 & 62.3$\pm$0.33 & 62.5$\pm$0.33 \\ \bottomrule
    \end{tabular}}
    \caption{Results of the Critical Question Answering task with and without Contextual Information [\textsc{ci}]. The reported results have been averaged from 3 randomly initialised sequential runs.}
    \label{tab:cqa}
\end{table}

We can observe a very different scenario when we look at the results on the CQA task. Comparing the results of the Binary Question Answering approaches involving fine-tuning (i.e., RoBERTa-\textsc{cqa}) with the Generative Question Answering using an LLM (i.e., Mixtral-8x7B-Instruct and GPT-4) without fine-tuning, the fine-tuned models were significantly better. Mixtral was the worst model for this task, followed by GPT-4. We observed no significant difference for the GPT-4 model with and without contextual information. In our experiments with RoBERTa-\textsc{cqa}, however, we were not only able to observe a  better performance in the CQA task compared to GPT-4 and Mixtral-8x7B, but the performance slightly yet significantly increased when considering the contextual information as part of the input. This is mostly due to the fact that external contextual information complements the information present in the argument with relevant details for answering the critical questions.


\subsection{Error Analysis}

\begin{figure}
    \centering
    \includegraphics[width=0.9\columnwidth]{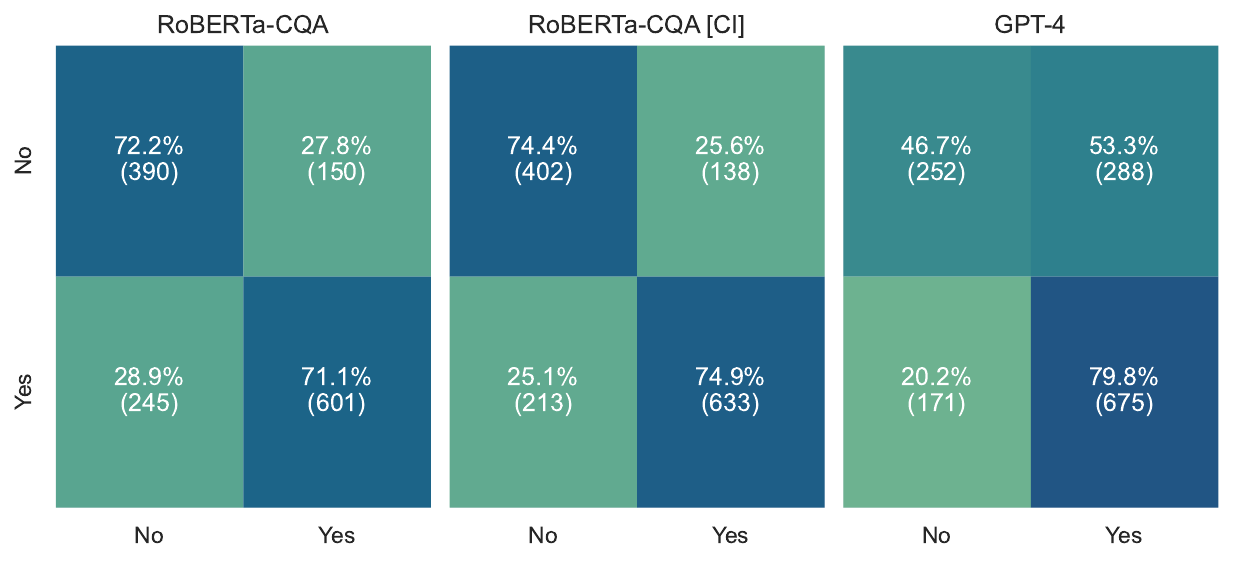}
    \caption{Aggregated confusion matrices of the CQA models (3 runs). Rows represent ground-truth values and columns the predicted answer to the CQs.}
    \label{fig:confus}
\end{figure}

\begin{figure*}
    \centering
    \includegraphics[width=0.7\textwidth]{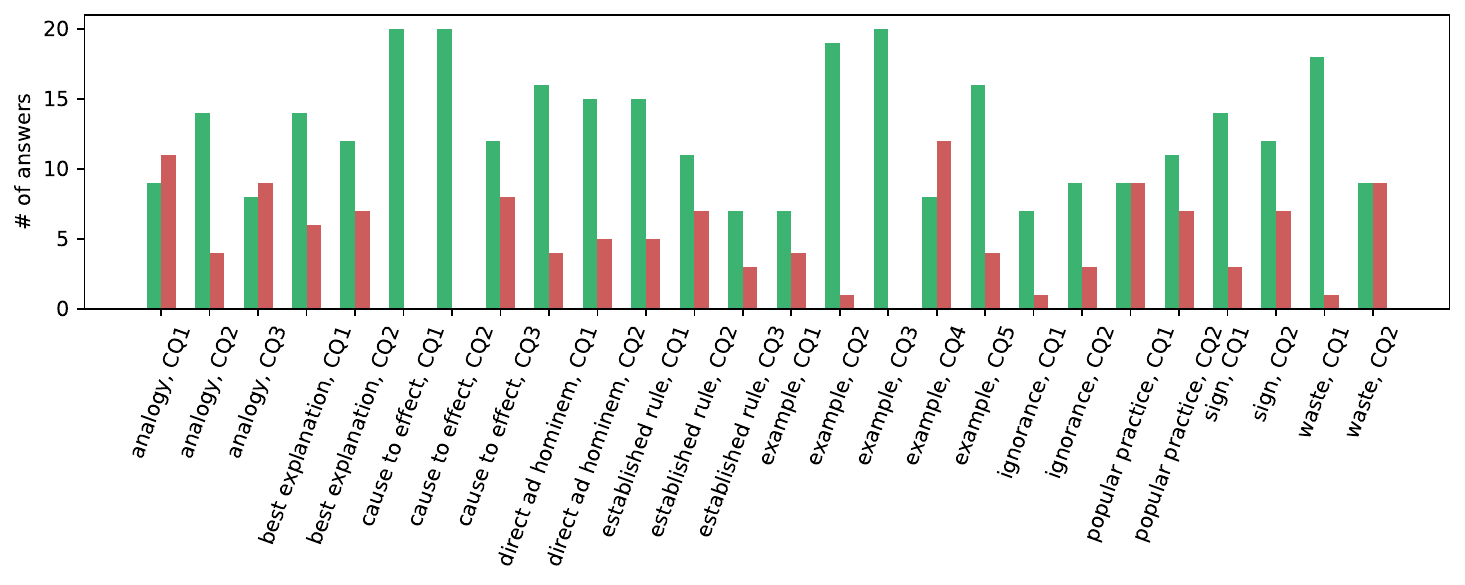}
    \caption{Distribution of the RoBERTa-\textsc{cqa} [\textsc{ci}] model outputs in the CQA test data. Green represents correct answers while red represents incorrect answers.}
    \label{fig:cqa_err}
\end{figure*}


In order to provide further details on the quantitative experimental results presented above, we conducted an analysis of the observed errors. Regarding ASC, we can clearly observe an important drop of performance when validating the models pre-trained on the NLAS-CQ corpus with natural language dialogue arguments. While models perform almost perfectly when classifying up to 20 different schemes considering textbook-like arguments where their structure is very clear and well defined, the task becomes significantly more challenging when doing so with incomplete arguments uttered in natural language dialogues. In dialogues, arguments do not have these perfectly defined structures, making it more difficult to establish boundaries between classes considering only the observable linguistic features. For example, an \textit{Argument from Waste} included in the NLAS-CQ corpus as:

\begin{itemize}[leftmargin=*]
    \item[] \underline{\textit{Premise 1}}: \textit{If someone focuses on their physical appearance and then suddenly stops trying to improve it, all their previous efforts will be wasted.}
    \item[] \underline{\textit{Premise 2}}: \textit{If all of someone's previous efforts to improve their physical appearance are wasted, their self-esteem and confidence may decrease, leading to a negative impact on their personal and professional success.}
    \item[] \underline{\textit{Conclusion}}: \textit{Therefore, it is important to continue trying to improve one's physical appearance in order to maintain and improve one's self-esteem and confidence, which can positively impact personal and professional success.} 
\end{itemize}

Is manifested very differently in the natural language dialogue validation data, consisting only of a premise and a conclusion (i.e., an enthymematic argument) as follows:

\begin{itemize}[leftmargin=*]
    \item[] \underline{\textit{Premise}}: \textit{We need to make sure that we \underline{embed} the \underline{successes} that we have had.}
    \item[] \underline{\textit{Conclusion}}: \textit{There is still work to do.}
\end{itemize}

Therefore, being able to determine the type of argumentation scheme falls into being able to correctly understand the meaning of \textit{embed} together with \textit{successes} within the given argumentative context, rather than being able to identify structural patterns of the natural language argument (a task in which language models stand out).

Secondly, we analyse the errors on CQA, the novel task proposed in this paper providing new dimensions to the previous work in misinformation identification. We excluded Mixtral-8x7B from this analysis due to its low performance on this task.

Looking at the confusion matrices in Figure \ref{fig:confus}, it is possible to observe some of the reasons behind the differences in performance reported in Table \ref{tab:cqa}. Our generative approaches present a tendency to accept most of the information as valid, resulting in a majority (53.3\%) of negative CQs answered affirmatively. This elevated percentage of errors represents an issue, making this approach far from usable. On the other hand, if we look at the fine-tuned binary QA models, we can observe a much more acceptable error distribution. Although there is still a moderate percentage of errors ($>25\%$) in both negative and affirmative answered CQs, these results represent a solid baseline to start researching in QA approaches for argumentative critical questions. Finally, it is also possible to observe the effect of including Contextual Information into the framework, with an improvement of more than a 2\% in the negative and almost a 5\% in the affirmative answered questions.

We can see a more detailed view of the results obtained by the RoBERTa-\textsc{cqa} [CI] model in Figure \ref{fig:cqa_err}. The distribution of the incorrectly answered CQs is not uniform, it is possible to see how it is dominated by some questions belonging to specific argumentation schemes. There are three specific cases where wrong answers outnumber correct answers: \textit{Argument from Analogy} CQ1 and CQ3, and \textit{Argument from Example} CQ4. The model also presents some limitations when answering CQs from \textit{Argument form Popular Practice} CQ1 and \textit{Argument from Waste} CQ2, where correct answers do not prevail over incorrect answers. Looking at these specific questions, it is possible to observe two potential reasons behind these errors. With regard to \textit{Argument from Analogy} CQ1 and CQ3, and \textit{Argument form Popular Practice} CQ1, providing a good answer to these questions depends largely on being able to retrieve high-quality contextual information supporting the answer. On the other hand, \textit{Argument from Example} CQ4 and \textit{Argument from Waste} CQ2 are highly subjective questions, one of the major challenges of the CQA task.

\section{Conclusion}

In this paper, we have explored misinformation identification from a new perspective, bridging the gap between argumentation theory and NLP, providing much-needed explainability. This paper represents a step forward from the state-of-the-art in misinformation detection, leaving behind binary (i.e., fallacy vs. non-fallacy) and multi class (i.e., fallacy type) sequence classifiers, and pointing towards a better informed, explainable approach. We are able to do this by combining concepts from argumentation theory (i.e., argumentation schemes and critical questions) with effective NLP algorithms. For that purpose, we create \textsc{NLAS-CQ}, the largest (and first) corpus of argumentation scheme instances along with answers to their corresponding critical questions. Using this corpus, we validate our proposed framework, showing how textbook-like arguments can be leveraged for ASC in natural dialogues, and providing a solid set of baselines for the CQA task. Future work includes improving the information retrieval process that complements our natural language input, and exploring effective ways of dealing with question subjectivity. 

\section*{Acknowledgements}

This work has been supported by the `AI for Citizen Intelligence Coaching against Disinformation (TITAN)' project, funded by the EU Horizon 2020 research and innovation programme under grant agreement 101070658, and by UK Research and innovation under the UK governments Horizon funding guarantee grant numbers 10040483 and 10055990.

\section*{Limitations}

We can identify two main limitations in our work. Although we report strong results in the Argumentation Scheme Classification task, these results correspond to the analysis of textbook-like natural language arguments. In human communication, we strangely make use of complete argument structures, it is common to use \textit{enthymemes}, shortening the arguments and making them feel more natural. Therefore, argumentative structures are used in a more heterogeneous way, making the Argumentation Scheme Classification task significantly harder as described in Table \ref{tab:validation}.

The second potential limitation of this work lies with the critical question answer distribution. Our corpus consists of 4,687 annotated critical questions which are distributed along 10 argumentation schemes and an average of 3 different questions per scheme, making it a very varied set in which each specific question is not evenly represented (see Appendix \ref{app:cqa_dist}). Extending the amount of available answered CQs might be as relevant as implementing an effective information retrieval module in order to significantly improve the reported results in the Critical Question Answering task.

\bibliography{custom}

\appendix

\section{Details of the \textsc{NLAS-CQ} Corpus}
\label{sec:appendix}
In this appendix, we provide specific details of the \textsc{NLAS-CQ} corpus, such as the argumentation schemes contained in it, the topics used to generate them, and descriptive statistics of the corpus per scheme and per topic.

\subsection{Argument Distribution per Argumentation Scheme}
\label{app:scheme_dist}

The argumentation schemes listed below were used for the creation of the \textsc{NLAS-CQ} corpus. Specific details about their structure and critical questions can be found in \cite{walton2008argumentation}. The number between parenthesis indicates the number of NLAS included in the corpus for each scheme. Marked with an asterisk ($*$) are the schemes whose critical questions have been annotated and included in the \textsc{NLAS-CQ} corpus.

\begin{itemize}
    \item Argument from Position to Know (197)
    \item Argument from Expert Opinion (199)
    \item ($*$)Direct ad Hominem (191)
    \begin{itemize}
        \item[-] CQ1: Is the allegation made in the attack to the character well supported by evidence?
        \item[-] CQ2: Is the issue of character relevant for the debate in which the argument was used?
    \end{itemize}
    \item Inconsistent Commitment (147)
    \item ($*$)Argument from Popular Practice (177)
    \begin{itemize}
        \item[-] CQ1: Are there actions or other indications that show that a large majority accepts the object of discussion?
        \item[-] CQ2: Even if a large majority accepts the object of discussion as true, are there grounds for thinking that they are justified in accepting the object of discussion?
    \end{itemize}
    \item Argument from Popular Opinion (183)
    \item ($*$)Argument from Analogy (194)
    \begin{itemize}
        \item[-] CQ1: Are there differences between the two cases that would tend to undermine the force of the similarity cited?
        \item[-] CQ2: Is the object of discussion true in the first case?
        \item[-] CQ3: Is there some other case that is also similar to the first case but in which the object of discussion is false?
    \end{itemize}
    \item Argument from Precedent (176)
    \item ($*$)Argument from Example (182)
    \begin{itemize}
        \item[-] CQ1: Is the proposition claimed in the premise in fact true?
        \item[-] CQ2: Does the example cited support the generalisation it is supposed to be an instance of?
        \item[-] CQ3: Is the example typical of the kinds of cases the generalisation covers?
        \item[-] CQ4: Is the generalisation strong?
        \item[-] CQ5: Do special circumstances of the example impair its generalisability?
    \end{itemize}
    \item ($*$)Argument from an Established Rule (168)
    \begin{itemize}
        \item[-] CQ1: Does the rule require carrying out types of actions that include the discussed action as an instance?
        \item[-] CQ2: Are there other established rules that might conflict or override this one?
        \item[-] CQ3: Is this case an exceptional one?
    \end{itemize}
    \item ($*$)Argument from Cause to Effect (191)
    \begin{itemize}
        \item[-] CQ1: Is the causal generalisation strong?
        \item[-] CQ2: Is the evidence cited strong enough to warrant the causal generalisation?
        \item[-] CQ3: Are there other causal factors that could interfere with the production of the effect in the given case?
    \end{itemize}
    \item Argument from Verbal Classification (185)
    \item Slippery Slope Argument (153)
    \item ($*$)Argument from Sign (166)
    \begin{itemize}
        \item[-] CQ1: Is the correlation of the sign with the event signified strong enough?
        \item[-] CQ2: Are there other events that would more reliably account for the sign?
    \end{itemize}
    \item ($*$)Argument from Ignorance (187)
    \begin{itemize}
        \item[-] CQ1: Has the search for evidence gone far enough?
        \item[-] CQ2: Is the proof supporting the claimed stance strong enough to be successful in fulfilling the burden?
    \end{itemize}
    \item Argument from Threat (152)
    \item ($*$)Argument from Waste (146)
    \begin{itemize}
        \item[-] CQ1: Is bringing about the mentioned proposition possible?
        \item[-] CQ2: Forgetting past losses that cannot be recouped, should a reassessment of the cost and benefits of trying to bring about the mentioned proposition from this point in time to be made?
    \end{itemize}
    \item Argument from Sunk Costs (183)
    \item Argument from Witness Testimony (198)
    \item ($*$)Argument from Best Explanation (192)
    \begin{itemize}
        \item[-] CQ1: Is the explanation satisfactory itself as an explanation of the finding?
        \item[-] CQ2: Is the explanation enough to draw a conclusion?
    \end{itemize}
\end{itemize}

\begin{figure*}
    \centering
    \includegraphics[width=\textwidth]{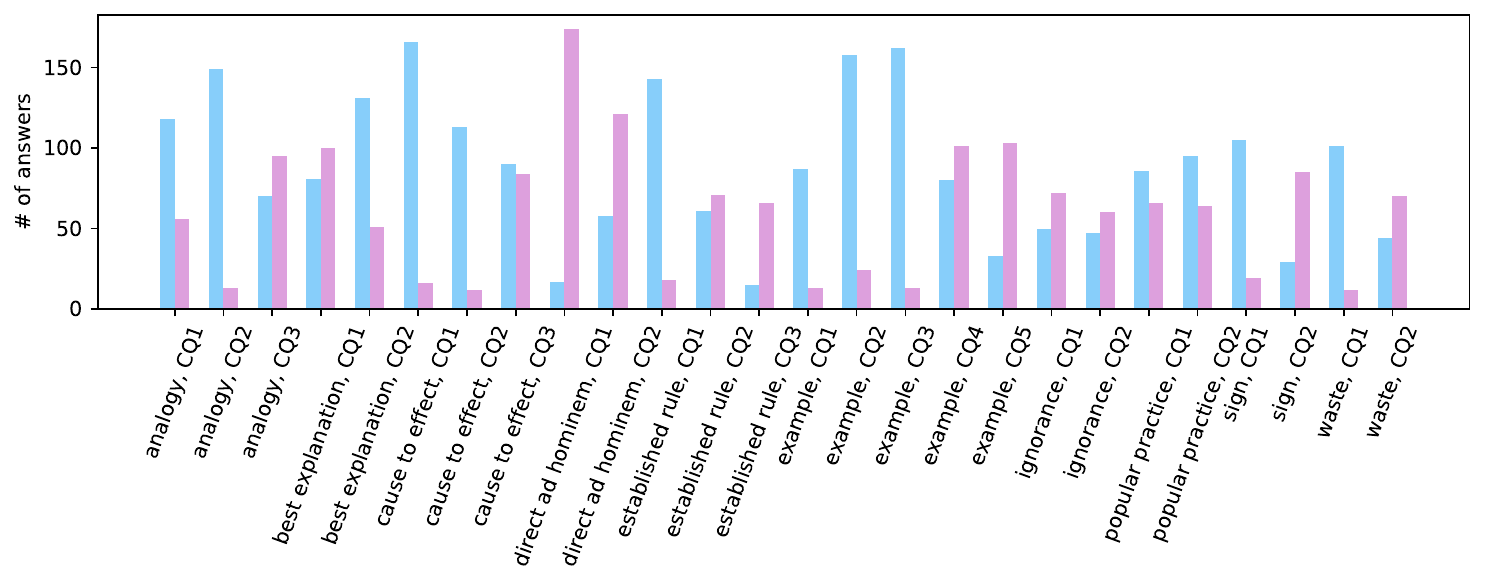}
    \caption{Distribution of the CQ answers in the complete \textsc{NLAS-CQ} corpus. Blue represents affirmative answers while pink presents negative answers.}
    \label{fig:cqa_dist}
\end{figure*}

\subsection{Argument Distribution per Topic}
\label{app:topics_dist}

For the generation of the \textsc{NLAS-CQ} corpus, we used a total of 100 topics. In the final corpus, the validated NLAS were distributed per topic as follows:

\textit{Euthanasia} (39), \textit{Mandatory vaccination in pandemic} (40), \textit{Physical appearance for personal success} (34), \textit{Intermittent fasting} (36), \textit{Capital punishment} (38), \textit{Animal testing} (37), \textit{Climate change} (39), \textit{Legalisation of cannabis} (37), \textit{Abortion} (39), \textit{Freedom of speech} (39), \textit{Tax increase} (38), \textit{Animal/human cloning} (36), \textit{Research in artificial intelligence} (38), \textit{Nuclear energy} (39), \textit{Use of online social networks} (39), \textit{Gun control} (38), \textit{Universal basic pension} (36), \textit{Gender quotas} (36), \textit{Genetic manipulation} (36), \textit{Reduction in working time} (39), \textit{Remote work} (39), \textit{Increasing security by sacrificing individual privacy} (38, \textit{Cryptocurrencies} (37), \textit{Censorship in social networks} (37), \textit{Terraplanism} (39), \textit{Renewable energy} (40), \textit{Electric transport} (38), \textit{Full self-driving cars} (38), \textit{Control measures to prevent economic inequality} (39), \textit{Immigration} (39), \textit{Offshore tax havens} (38), \textit{Tariffs on imported products} (40), \textit{Assisted suicide} (39), \textit{Birth control} (39), \textit{Globalisation} (39), \textit{Internet censorship} (39), \textit{Legalisation of prostitution} (37), \textit{Use of nuclear weapons} (39), \textit{Immortality} (37), \textit{Surrogacy} (37), \textit{Indiscriminate launching of satellites} (36), \textit{Drone strikes} (36), \textit{Internet access for children} (38), \textit{School uniform} (37), \textit{Regulation of unhealthy foods} (38), \textit{Political correctness} (39), \textit{UFO existence} (36), \textit{Chemtrail conspiracy theory} (35), \textit{Use of masks in public spaces} (39), \textit{Sustainable Development Goals} (38), \textit{Right of self-determination} (35), \textit{International economic sanctions} (33), \textit{Animal rights} (38), \textit{Outsourcing and privatization} (38), \textit{Ban of Burka in public spaces} (39), \textit{Gender-based violence} (29), \textit{Universal basic income} (38), \textit{Homeschooling} (38), \textit{Artificial intelligence in warfare} (36), \textit{Use of recreational drugs} (35), \textit{Universal suffrage} (36), \textit{Reducing the use of plastic} (38), \textit{Space exploration} (34), \textit{Enlargement of NATO} (32), \textit{Violent video game restriction} (31), \textit{Positive discrimination} (30), \textit{International border opening} (31), \textit{Religious education in public schools} (31), \textit{Sex education} (30), \textit{Use of chemical pesticides} (29), \textit{Compulsory military service} (34), \textit{Space tourism} (31), \textit{Private funding of political campaigns} (34), \textit{Ethics control in scientific research} (33), \textit{Free public Wi-Fi access points} (35), \textit{Regulation of streaming services} (37), \textit{Facial recognition technology in public spaces} (38), \textit{Online voting} (38), \textit{Social reintegration} (36), \textit{Automated fake news detection systems} (36), \textit{Bias in AI algorithms} (32), \textit{Planned obsolescence} (34), \textit{Degrowth economics} (38), \textit{Investing in public education} (37), \textit{Military presence in public places} (38), \textit{Capitalism} (36), \textit{Communism} (37), \textit{Anarchism} (33), \textit{Nationalisation of the banks} (33), \textit{Liberalism} (29), \textit{Bullfighting} (32), \textit{Fishing size and catch limits} (26), \textit{Hunting restrictions} (28), \textit{Housing speculation} (26), \textit{Government intervention} (25), \textit{Institutional support for the LGBTI community} (32), \textit{Protection of minority languages} (29), \textit{Ban of zoos} (35), \textit{Online piracy} (30), and \textit{Open source research} (31).

\subsection{Critical Question Answer Distribution}
\label{app:cqa_dist}

Figure \ref{fig:cqa_dist} depicts the distribution of affirmative and negative answers to the complete set of annotated critical question for 10 argumentation schemes in a range of one hundred different topics.

\subsection{Contextual Information}
\label{app:context}

Regarding the contextual information, we asked our annotators to support the answers provided to the CQs. The information retrieval process, however, falls out of the scope of this paper, assuming that this information is available without the needs of automatically obtaining it, to prove its usefulness in the CQA task. Therefore, some examples of the contextual information used in our experiments include:

\begin{itemize}
    \item Argument from Best Explanation (Terraplanism)
        \begin{itemize}
            \item[-] CQ1: ``\textit{The globe has been understood for thousands of years. This was one of the first cosmic facts to be worked out correctly by ancient people because evidence of a spherical Earth is visible to the naked eye. Sailors noticed that the sails of approaching ships appeared before the hulls of the ships became visible, because the surface of Earth is slightly curved like the surface of an enormous ball.}''
            \item[-] CQ2: ``\textit{Like the other planets, the Earth is a sphere. It rotates around its own axis slowly but continuously, and completes this rotation about once every 24 hours. As the Earth rotates, part of the Earth is facing the sun, and part of it faces away. This is why some parts of the world have day and night at different times.}''
        \end{itemize}
    \item Argument from Popular Practice (Cryptocurrencies)
        \begin{itemize}
            \item[-] CQ1: ``\textit{The committee said "unbacked" crypto assets - typically cryptocurrencies with no fixed value - exposed "consumers to the potential for substantial gains or losses, while serving no useful social purpose". "These characteristics more closely resemble gambling than a financial service," the MPs added}''
            \item[-] CQ2: ``\textit{MPs have urged the government to treat retail investment in cryptocurrencies such as Bitcoin as a form of gambling}''
        \end{itemize}
\end{itemize}

\subsection{Prompts for NLAS Generation}
\label{app:prompt}

For generating the 3,566 NLAS we employed the following prompt structure. Variables such as \{\textit{scheme}\}, \{\textit{stance}\}, \{\textit{topic}\}, or \{\textit{argument}\_\textit{description}\} were adapted each time to obtain the variety and richness described in the paper.

\begin{itemize}
    \item[] ``Provide an \{\textit{scheme}\} argument \{\textit{stance}\} on the topic of \{\textit{topic}\} following the argument pattern:
    
    \{\textit{argument}\_\textit{description}\}
    
    You must answer in a json format following the structure of the argumentation scheme.'' 
\end{itemize}

The first variable, \{\textit{scheme}\}, was replaced by the name of one of the generated argumentation schemes. \{\textit{stance}\} was replaced by ``\textit{in favour}'' or ``\textit{against}''. The variable \{\textit{topic}\} was instantiated as one of the 100 topics included in the list above. Finally, the \{\textit{argument}\_\textit{description}\} was replaced by the argumentation scheme structure as defined in \cite{walton2008argumentation}.

\subsection{Prompt for Critical Question Answering}
\label{app:cqa-prompt}

Both Mixtral-8x-7B-Instruct-v0.1 and GPT-4 were prompted using the same prompt for addressing the CQA task. The complete prompt used in this process was defined as follows,

\begin{itemize}
    \item[] ``You are a binary question answering system. Reply [yes] or [no] to the following question: 

    context: \{\textit{context}\}

    question: \{\textit{question}\}
    
    answer: '' 
\end{itemize}

where \{\textit{context}\} was completed with the argumentation scheme instances and the contextual information (if included in the experiment), and \{\textit{question}\} was replaced by each critical question related to the argumentation scheme. This way, if a given argumentation scheme had multiple CQs, different prompts were generated for each question so that the model could answer them individually.

\end{document}